\documentclass[journal,12pt,onecolumn,draftclsnofoot,]{IEEEtran}
\usepackage{cite}

\ifCLASSINFOpdf
   \usepackage[pdftex]{graphicx}
\else
\fi
\usepackage{amsmath}

\usepackage{xcolor}

\begin{document}

\title{Probabilistic Deterministic Finite Automata and\\
Recurrent Networks, Revisited}


\author{Sarah E. Marzen and~James~P.~Crutchfield
\thanks{SEM is with W. M. Keck Science Department,
Claremont McKenna, Scripps, and Pitzer College, 925 N Mills Ave, Claremont, CA 91711 
USA; e-mail: smarzen@cmc.edu.}
\thanks{JPC directs the Complexity Sciences Center, Physics
Department, University of California, Davis CS 95616; email: chaos@ucdavis.edu.}
\thanks{Manuscript received October 1, 2019.}}

\markboth{IEEE JSAIT Inaugural Issue 2019 ~~~~~ Deep Learning: Mathematical
Found'ns
Appl'ns to Info. Sci.}%
{Shell \MakeLowercase{\textit{et al.}}: Bare Demo of IEEEtran.cls for IEEE Journals}

\maketitle

\begin{abstract}
Reservoir computers (RCs) and recurrent neural networks (RNNs) can mimic any
finite-state automaton in theory, and some workers demonstrated that this can
hold in practice. We test the capability of generalized linear models,
RCs, and Long Short-Term Memory (LSTM) RNN architectures to
predict the stochastic processes generated by a large suite of probabilistic
deterministic finite-state automata (PDFA). PDFAs provide an excellent
performance benchmark in that they can be systematically enumerated, the
randomness and correlation structure of their generated processes are exactly
known, and their optimal memory-limited predictors are easily computed.
Unsurprisingly, LSTMs outperform RCs, which outperform generalized
linear models. Surprisingly, each of these methods can fall short of the
maximal predictive accuracy by as much as $50\%$ after training and, when optimized, tend to
fall short of the maximal predictive accuracy by $\sim 5 \%$, even though
previously available methods achieve maximal predictive accuracy with
orders-of-magnitude less data. Thus, despite the representational universality
of RCs and RNNs, using them can engender a surprising predictive gap for simple
stimuli. One concludes that there is an important and underappreciated role for
methods that infer ``causal states'' or ``predictive state representations''.
\end{abstract}


\IEEEpeerreviewmaketitle

\section{Introduction}
%
%
%
%

\IEEEPARstart{M}{any} seminal results established that both reservoir computers
(RCs) \cite{maass2002real,grigoryeva2018echo} and recurrent neural networks
(RNNs) \cite{doya1993universality} can reproduce any dynamical system, when
given a sufficient number of nodes. Further work gave example RNNs that
faithfully reproduce finite state automata, to the point that RNN nodes
mimicked the automata states \cite{cleeremans1989finite}, and established
bounds on the required RNN complexity \cite{horne1994bounds}. One would
conjecture, then, that Long Short-Term Memory (LSTM) architectures---an easily
trainable RNN variety \cite{schmidhuber1997long, collins2016capacity}---should
easily learn to predict the outputs of probabilistic deterministic finite
automata (PDFA), also called unifilar hidden Markov models in information
theory. The PDFAs used in the following are simple, in that their statistical
complexity \cite{Shal98a} and excess entropy \cite{Bial00a, Crut01a} are finite
and relatively small. The following explores PDFAs since optimal predictors of
the time series they generate are easily computed \cite{Shal98a}, and the
tradeoffs between code rate and predictive accuracy (encapsulated by the
predictive rate-distortion function) are easily computed as well \cite{Marz14f}.

We use predictive rate-distortion functions to calibrate the performance of
three time series predictors: generalized linear models (GLMs)
\cite{nelder1972generalized}, RCs \cite{maass2002real,grigoryeva2018echo}, and
LSTMs \cite{schmidhuber1997long}. Unsurprisingly, LSTMs are generally more
accurate and efficient than reservoirs, which are generally more accurate and
efficient than GLMs. Surprisingly, despite the simplicity of the generated
stochastic time series, we find that all tested prediction methods fail to
attain maximal predictive accuracy by as much as $50\%$ and often need higher
rates than necessary to attain that predictive performance. However, existing
methods for inferring PDFAs \cite{Stre13a} can correctly infer the PDFA and
generate the optimal predictor with orders-of-magnitude less data. This leads
us to conclude that prediction algorithms that first infer \emph{causal states}
\cite{Crut88a, Stre13a, pfau2010probabilistic, still2014information} can
surpass trained RNNs if the time series in question has (approximately) finite
causal states. (Causal states are sometimes also called \emph{predictive state
representations} \cite{littman2002predictive}.)

\section{Using predictive rate-distortion to evaluate time series prediction algorithms}
\label{sec:PRD}

Many real-world tasks rely on prediction. Given past stock prices, traders try
to predict if a stock price will go up or down, adjusting investment strategies
accordingly. Given past weather, farmers endeavor to predict future
temperatures, rainfall, and humidity, adapting crop and pesticide choices.
Manufacturers try to predict which goods will appeal most to consumers,
adjusting raw materials purchases. Self-driving cars must predict the motion of
other objects on and off the road. And, when it comes to biology, evidence
suggests that organisms endeavor to predict their environment as a key survival
strategy \cite{schultz1997neural, montague1996framework, rao1999predictive}.

However, we also care about the cost of communicating a prediction, either to
another person or from one part an organism to another. Channel capacity can be
energetically expensive. All other concerns equal, one is inclined to employ a
predictor with a lower transmission rate \cite{Berg71a}.

Simultaneously optimizing the objectives---high predictive accuracy and low
code rate---leads to \emph{predictive rate-distortion} \cite{still2010optimal,
still2014information, Marz14f}. With an eye to making contact with
nonpredictive rate-distortion theory, we summarize the setup of predictive
rate-distortion as follows. Semi-infinite pasts are sent i.i.d. to an encoder,
which then produces a prediction or a probability distribution over possible
predictions. The predictive distortion measures how far the estimated
predictions differ from correct predictions. Distortion is often taken, for
example, to be the Kullback-Leibler divergence between the true distribution
$p(\overrightarrow{x}|\overleftarrow{x})$ over futures $\overrightarrow{x}$
conditioned on the past $\overleftarrow{x}$ and the distribution
$p(\overrightarrow{x}|r)$ over futures conditioned on our \emph{representation}
$r$ \cite{tishby2000information}. The predictive rate-distortion function
$R(D)$ separates the plane of rates and predictive distortions into regions of
achievable and unachievable combinations. A slight variant of the
rate-distortion theorem gives:
\begin{align}
R(D) = \min_{p(\vec{x}|r): E[d] \leq D} I[\overleftarrow{X};R]
  ~,
\label{eq:PRD}
\end{align}
where $I[\cdot;\cdot]$ is the mutual information. When the distortion is the
Kullback-Leibler divergence, the predictive rate-distortion function is
directly related to the predictive information curve \cite{still2010optimal,
still2014information}. Finding representations that lie on the rate-distortion
curve motivates slow feature analysis \cite{creutzig2008predictive}, recovers
canonical correlation analysis \cite{creutzig2009past}, and identifies the
minimal sufficient statistics of prediction---the causal states
\cite{still2010optimal}. Predictive information curves have even been used to
evaluate the predictive efficiency of salamander retinal neural spiking
patterns \cite{palmer2015predictive}.

Here, however, we adopt the stance that predictive accuracy---the probability
that one's prediction is correct---is more natural than a Kullback-Leibler
divergence. Accordingly, we force our representation $r\in\{0,1\}$ to be a
prediction, and calculate \emph{accuracy} via the distortion measure:
\begin{align*}
d(r_t,x_{t+1}) = \delta_{r_t,x_{t+1}}
  ~,
\end{align*}
which implies:
\begin{align*}
E[d] = \sum_{\overleftarrow{x}_t} p(\overleftarrow{x}_t) \sum_{r_t = x_{t+1}}  p(r_t|\overleftarrow{x}_t) p(x_{t+1}|\overleftarrow{x}_t)
  ~.
\end{align*}
The corresponding predictive rate-accuracy function is almost as in Eq.
(\ref{eq:PRD}), except with a changed constraint:
\begin{align}
R(A) = \min_{p(\vec{x}|r): E[d] \geq A} I[\overleftarrow{X};R].
\label{eq:PRA}
\end{align}
This is closer in spirit to the information curve than the rate-distortion
function, in that the achievable region lies below the predictive rate-accuracy
function.

\section{Background}

In what follows, we review time-series generation and the widely-used
prediction methods we compare. We first discuss PDFAs and then prediction
methods.

\subsection{PDFAs and predictive rate-distortion}

We focus on minimal PDFAs---for a given stochastic process that with the
smallest number of states. A PDFA consists of a set $\mathcal{S}$ of
states $\sigma\in\mathcal{S}$, a set $\mathcal{A}$ of emission symbols, and
transition probabilities $p(\sigma_{t+1},x_t|\sigma_t)$, where $\sigma_t$,
$\sigma_{t+1} \in\mathcal{S}$ and $x_t\in\mathcal{A}$. The ``deterministic''
descriptor comes from the fact that $p(\sigma_{t+1}|x_t,\sigma_t)$ has support
on only one state. (This is ``determinism'' in the sense of formal language
theory \cite{Hopc79}---an automaton deterministically \emph{recognizes} a
string---not in the sense of nonstochastic. It was originally called
\emph{unifilarity} in the information theoretic analysis of hidden Markov
chains \cite{Ash65a}. Thus, PDFAs are also known as \emph{unifilar hidden
Markov models} \cite{Shal98a}.)

Here, we concern ourselves with minimal and binary-alphabet ($\mathcal{A} =
\{0,1\}$) PDFAs. In dynamical systems theory minimal unifilar HMMs (minimal
PDFAs) are called \emph{$\epsilon$-machines} and their states $\sigma$
\emph{causal states}. Due to the automaton's determinism, one can uniquely
determine the state from the past symbols with probability $1$. Each state is
therefore a cluster of pasts that have the same conditional probability
distribution over futures. As a result, all that one needs to know to
optimally predict the future is given by the causal state \cite{Shal98a}.

\begin{figure}[!t]
\centering
\includegraphics[width=2.5in]{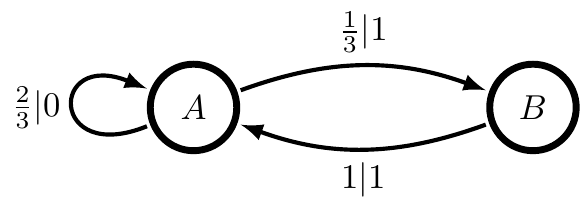}
\caption{Minimal two-state PDFA that generates the Even Process, so-called
	since there are always an even number of $1$s between $0$'s. Arrows
	indicate allowed transitions, while transition labels $p|s$ indicate the
	transition (and so too emission) probabilities $p \in [0,1]$ for the symbol
	$s \in \mathcal{A}$. Given a current state and next symbol, one knows the
	next state---the deterministic or unifilar property of this PDFA.
	}
\label{fig:EvenProcess}
\end{figure}

For example, the simple two-state PDFA shown in Fig. \ref{fig:EvenProcess}
generates the Even Process: only an even number of $1$'s are seen between two
successive $0$'s. This leads to a simple prediction algorithm: find the parity
of the number of $1$'s since the last $0$; if even, we are in state $A$, so
predict $0$ and $1$ with equal probability; if odd, we are in state $B$, so
predict $1$. There is only one past for which our prediction algorithm yields
no fruit: given the past of all $1$s a single state is never identified. One
only knows that the machine is in either state $A$ or $B$ and the best
prediction is a mixture of what the states indicate. Even though that
past occurs with probability $0$, it causes the Even Process to be an
infinite-order Markov Process \cite{Jame10a}. See Ref. \cite{lohrthesis} for a
measure-theoretic treatment.

Causal states and $\epsilon$-machines can be inferred from data in a variety of ways \cite{shalizi2002pattern, Stre13a, pfau2010probabilistic, still2014information}.

The causal states are uniquely useful to calculating predictive rate-distortion
curves. Under weak assumptions, the predictive rate-accuracy function of
Sec. \ref{sec:PRD} becomes:
\begin{align*}
R(A) =  \min_{p(r|\sigma): E[d] \geq A} I[\mathcal{S};R]
\end{align*}
with:
\begin{align*}
E[d] = \sum_{\sigma_t} p(\sigma_t) \sum_{x_{t+1}=r_t} p(r_t|\sigma_t) p(x_{t+1}|\sigma_t)
  ~.
\end{align*}
See Ref. \cite{Marz14f} for the proof. With this substitution---of a finite
object ($\mathcal{S}$) for an infinite one ($\overleftarrow{X}$)---the
Blahut-Arimoto algorithm can be used to accurately calculate the predictive
rate-accuracy function, in that the algorithm provably converges to the optimal
$p(r|\sigma)$ \cite{csiszar1974computation}. The same cannot be said of the
predictive information curve \cite{Marz14f}, which converges to a local optimum
of the objective function, but may not converge to a global optimum.

In practice, we always augment the predictive rate-accuracy function with the
rate and accuracy of the optimal predictor, which is (as described earlier)
straightforwardly derived from the $\epsilon$-machine. Simply put, we infer the
causal state $\sigma_t$ from past data and predict the next symbol to be
$\arg\max_{x_{t+1}} p(x_{t+1}|\sigma_t)$.

The following tests the various time series predictors on all of the (uniformly
sampled) binary-alphabet $\epsilon$-machine topologies \cite{John10a} with
randomly-chosen emission probabilities. Due to the super-exponential explosion
of the set of topological $\epsilon$-machines with number of states, we only
look at binary-alphabet machines with four or fewer (causal) states. (There are
$1,338$ unique topologies for four states, but over $10^6$ for six states.) The
analysis discards any $\epsilon$-machine with zero-rate optimal predictor,
which can arise depending on the emission probabilities.

\subsection{Time series methods}

We focus on three methods for time series prediction: generalized linear models
(GLM), reservoir computers (RCs), and LSTMs.

The GLM we use predicts $x_t$ from a linear combination of the last $k$ symbols
$x_{t-k},x_{t-k+1},...,x_{t-1}$. More precisely, a GLM models the probability
of $x_t$ being a $0$ via:
\begin{align}
p_{GLM}(x_t=0|x_{t-k},...,x_{t-1}) = \frac{e^{w_k x_{t-k} + ... + w_1 x_{t-1} + w_0}}{1+e^{w_k x_{t-k} + ... + w_1 x_{t-1} + w_0}}
  ~.
\label{eq:pGLM0}
\end{align}
The model's estimate of the probability of $x_t=1$ follows:
\begin{equation}
p_{GLM}(x_t=1|x_{t-k},...,x_{t-1}) = \frac{1}{1+e^{w_k x_{t-k} + ... + w_1 x_{t-1} + w_0}}
  ~.
\label{eq:pGLM1}
\end{equation}
We use Scikit-learn logistic regression to find the best weights
$w_0,w_1,...,w_k$. Predictions are then made via $\arg\max_{x_t} p_{GLM}(x_t|x_{t-k},...,x_{t-1})$.

The RC is more powerful in that it uses logistic regression with features that
contain information about symbols arbitrarily far into the past. We employ a
$tanh$ activation function, so that the reservoir's state advances via:
\begin{align}
h_{t+1} = \tanh (W h_t + v x_t + b)
\label{eq:res}
\end{align}
and initialize $W,~v,~b$ with i.i.d. normally distributed elements. The matrix
$W$ is then scaled so that it is near the ``edge of chaos'' \cite{Crut89e,
Pack88a, Mitc93b, Mitc93a}, where RCs are conjectured to have maximal memory
\cite{bertschinger2004real,boedecker2012information}. We then use logistic
regression with $h_t$ as features to predict $x_t$:
\begin{align*}
p_{reservoir}(x_t = 0 | h_t) &= \frac{e^{w^{\top}h_t + w_0}}{1+e^{w^{\top}h_t +
w_0}} ~, \\
p_{reservoir}(x_t = 1 | h_t) &= \frac{1}{1+e^{w^{\top}h_t + w_0}}
  ~.
\end{align*}
It is straightforward to devise a weight matrix $W$ and bias $b$ so that
$p_{reservoir}(x_t|h_t)$ attains the restricted linear form of $p_{GLM}$ of
Eqs. (\ref{eq:pGLM0}) and (\ref{eq:pGLM1}). That is, RCs are more
powerful than GLMs. We use Scikit-learn logistic regression to find the best
weights $w_0$ and $w$. Note that the weights $W$, $v$, and $b$ are not learned,
but held constant; we only train $w$ and $w_0$. Predictions are made via
$\arg\max_{x_t} p_{reservoir}(x_t|h_t)$.

Finally, we analyze the LSTM's predictive capabilities. LSTMs are no more
powerful than vanilla RNNs; e.g., those like in Eq. (\ref{eq:res}). However,
they are far more trainable in that it is possible to achieve good results
without extensive hyperparameter tuning \cite{collins2016capacity}. An LSTM has
several hidden states $f_t$, $i_t$, $o_t$, $c_t$, and $h_t$ that update via the following:
\begin{align*}
f_t &= \sigma_g (W_f x_t + U_f h_{t-1}+b_f) \\
i_t &= \sigma_g (W_i x_t + U_i h_{t-1}+b_i) \\
o_t &= \sigma_g (W_o x_t + U_o h_{t-1}+b_o) \\
c_t &= f_t \odot c_{t-1} + i_t \odot \sigma_c (W_c x_t + U_c h_{t-1} + b_c) \\
h_t &= o_t \odot c_t
  ~,
\end{align*}
where $\sigma_g$ is the sigmoid function and $\sigma_c$ is the hyperbolic
tangent.  The variable $c_t$ is updated linearly, therefore avoiding issues
with vanishing gradients \cite{hochreiter1998vanishing}.  Meanwhile, the gating
function $f_t$ allows us to forget the past selectively.  We then predict the
probability of $x_t$ given the past using:
\begin{align}
p_{LSTM}(x_t = 0 | h_t) &= \frac{e^{w^{\top}h_t + w_0}}{1+e^{w^{\top}h_t +
w_0}} ~, \nonumber \\
p_{LSTM}(x_t = 1 | h_t) &= \frac{1}{1+e^{w^{\top}h_t + w_0}}
  ~.
\label{eq:LSTMprobs}
\end{align}
Weights $w$ and $w_0$ are learned while we estimate parameters $W_f$, $U_f$,
$b_f$, $W_i$, $U_i$, $W_o$, $U_o$, $b_o$, $W_c$, $U_c$, and $b_c$ to maximize
the log-likelihood. Predictions are made via $\arg\max_{x_t} p_{LSTM}(x_t|h_t)$.

Predictive accuracy is calculated by comparing the predictions to the actual
values of the next symbol and counting the frequency of correct predictions.
The code rate is calculated via the prediction entropy \cite{Berg71a}.

\section{Results}

Our aim here is to thoroughly and systematically analyze the predictive
accuracy and code rate of our three time series predictors of a large swath of
PDFAs. To implement this, we ran through Ref. \cite{John10a}'s
$\epsilon$-machine library---binary-alphabet PDFAs with four states or less and
randomly chosen emission probabilities. For each PDFA, we generated a
length-$5000$ time series. The first half was presented to a predictor and used
to train its weights. We then evaluated each time series predictor based on its
predictions for the second half of the time series. Predictive accuracy and
code rate were calculated and compared to the predictive rate-distortion function.

Note that Bayesian structural inference (BSI) provides a useful comparison
\cite{Stre13a}. In BSI, we compute the maximum a posteriori (MAP) estimate of
the PDFA generating an observed time series, and use this MAP estimate to build
an optimal predictor of the process. BSI can correctly infer the PDFA
essentially $100\%$ of the time with orders-of-magnitude less data than used to
monitor the three prediction methods tested here. Hence, it achieves optimal
predictive accuracy with minimal rate. Our aim is to test the ability of GLMs,
RCs, and RNNs to equal BSI's previously-published performance.


The time series predictors used have hyperparameters. A variety of orders
($k$'s) were used for the GLMs and reservoirs and LSTMs of different sizes
(number of nodes) were tested. Learning rate and optimizer type, including
gradient descent and Adam \cite{kingma2014adam}, were also varied for the LSTM,
with little effect on results.

For the most part, we find that all three prediction methods--GLMs, RCs, and
LSTMs---learn to predict the PDFA outputs near-optimally, in that prediction
accuracies differ from the optimal prediction accuracy by an average of roughly
$5\%$. LSTMs outperform RCs, which outperform GLMs. However, we discovered
simple PDFAs that cause the best LSTM to fail by as much as $5\%$, the best RC
to fail by as much as $10\%$, and the best GLM to fail by as much as $27\%$.

This leads us to conclude that existing methods for inferring causal states
\cite{shalizi2002pattern, Stre13a, pfau2010probabilistic, still2014information}
are useful, despite the historically dominant reliance on RNNs. For example,
as previously mentioned, Bayesian structural inference correctly infers the
correct PDFAs almost $100\%$ of the time, leading to essentially zero
prediction error, on training sets that are orders of magnitude smaller than
those used here \cite{Stre13a}.

\subsection{The difference between theory and practice: the Even and Neven Process}

We first analyze two easily-described PDFAs, deriving RNNs that correctly infer
causal states and, therefore, that match the optimal predictor---the
$\epsilon$-machine. We then compare the trained GLMs, RCs, and LSTMs to the
easily-inferred optimal predictors. In theory, RCs and LSTMs should be able to
mimic the derived RNNs, in that it is possible to find weights of an RC and
LSTM that yield nodes that mimic the causal states of the PDFA. In practice,
surprisingly, RCs and LSTMs have some difficulty.

\begin{figure}[!t]
\centering
\includegraphics[width=2.5in]{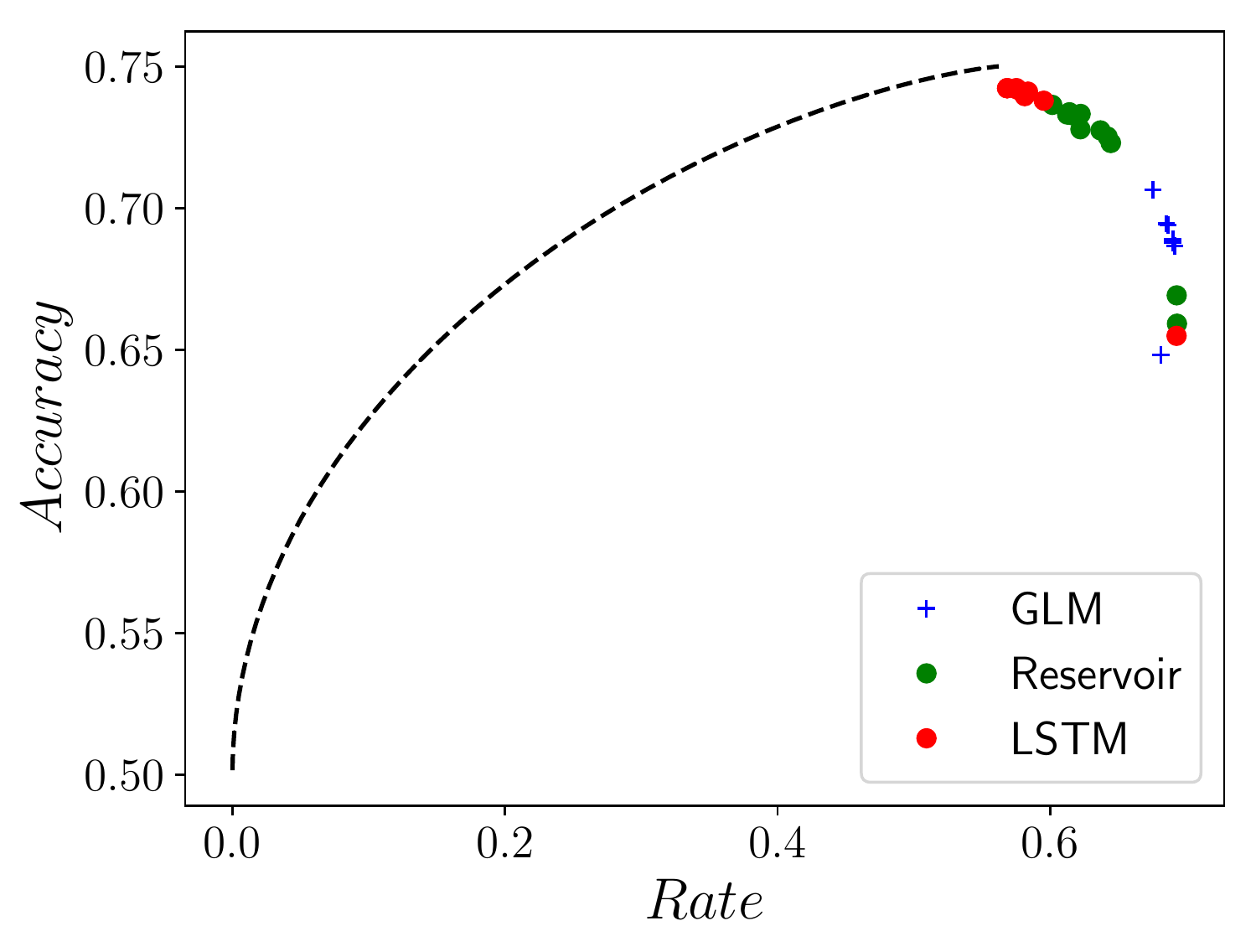}
\caption{Predictive rate--accuracy curve for the Even Process in Fig.
	\ref{fig:EvenProcess}, along with empirical predictive accuracies and rates
	of GLMs, RCs, and LSTMs of various sizes: orders range from $1$-$10$ for
	GLMs, number of nodes range from $1$-$61$ for RCs, and number of nodes
	range from $1$-$121$ for LSTMs.  Despite the Even Process' simplicity,
	there is a noticeable difference between the predictors' performances and
	the optimal achievable performance.
	}
\label{fig:EvenProcessPRD}
\end{figure}

First, we analyze the Even Process shown in Fig. \ref{fig:EvenProcess}. The
optimal prediction algorithm is easily seen by inspection of Fig.
\ref{fig:EvenProcess}. When we determine the machine is in state $A$, we
predict a $0$ or a $1$ with equal probability; if it is in state $B$, we
predict a $1$. We determine whether or not it is in state $A$ or $B$ by the
parity of the number of $1$s since the last $0$. If odd, it is in state $B$; if
even, it is in state $A$. The inferred state is easily encoded by the following
RNN:
\begin{equation}
h_{t+1} = x_t(1-h_t)
  ~.
\label{eq:RNN_Even}
\end{equation}
If $x_t$ is $0$, the hidden state of the RNN ``resets'' to $0$; e.g., state
$A$. If $x_t=1$, then the hidden state updates by flipping from $0$ to $1$ or
vice versa, mimicking the transitions from $A$ to $B$ and back. One can show
that a one-node LSTM hidden state $h_t$ can, with proper weight choices, mimic
the hidden state of Eq. (\ref{eq:RNN_Even}). With the correct hidden state
inferred, it is straightforward to find $w$ and $w_0$ such that Eq.
(\ref{eq:LSTMprobs}) yields optimal (and correct) predictions.

As one might then expect, and as Fig. \ref{fig:EvenProcessPRD} confirms, LSTMs
tend to have rates that are close to the optimal (maximal) rate and predictive
accuracies that are only slightly below the optimal predictive accuracy. RCs
and GLMs tend to have higher rates and lower predictive accuracies, but they
are still within $\sim 13\%$ of optimal. As one might also expect, LSTMs and
RCs with additional nodes and GLMs with higher orders (higher $k$) have higher
predictive accuracies than LSTMs and RCs with fewer nodes and GLMs with lower
orders. But viewed another way, given the simplicity of the stimulus---indeed,
given that a one-node LSTM can, in theory, learn the Even Process---the gap
from the predictors' rates and accuracies to the optimal combinations of rate
and accuracy is surprising. It is also surprising that none of the three
predictors' rates fall below the maximal optimal rate.

\begin{figure}[!t]
\centering
\includegraphics[width=2.5in]{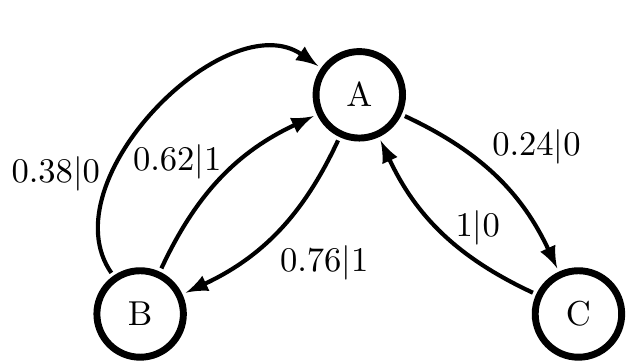}
\includegraphics[width=2.5in]{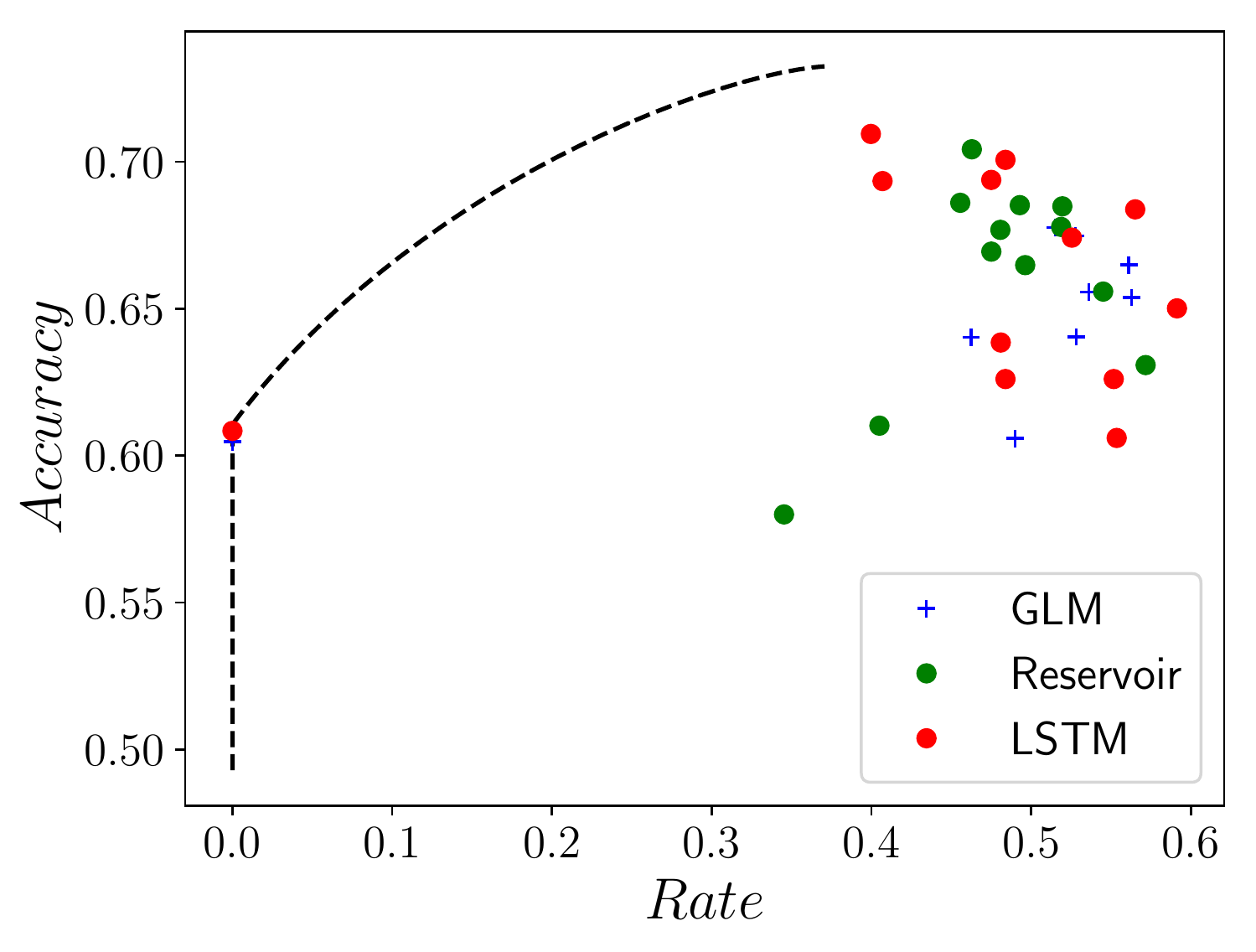}
\caption{Predictive rate-accuracy curve for the Neven Process (PDFA shown at
	top), along with empirical predictive accuracies and rates of GLMs, RCs,
	and LSTMs of various sizes: orders range from $1$-$10$ for GLMs, number of
	nodes range from $1$-$61$ for RCs, and number of nodes range from $1$-$121$
	for LSTMs. Despite Neven Process' simplicity, there is a noticeable gap
	between the predictor's performance and the optimal performance
	achievable.
	}
\label{fig:Neven}
\end{figure}

Figure \ref{fig:Neven} introduces a similarly-simple three-state PDFA. If a
$1$ is observed after a $0$, we are certain the machine is in state $B$; after
state $B$, we know it will transition to state $A$; and then the parity of $0$s
following transition to state $A$ tells us if it is in state $A$ (even) or
state $B$ (odd). This PDFA is a combination of a Noisy Period-$2$ Process
(between states $A$ and $B$) and an Even Process (between states $A$ and $C$).

Given the Neven Process's simplicity, it is unsurprising that we can concoct an
RNN that can infer the internal state. Let $h_t = (h_{t,A},h_{t,B},h_{t,C})$ be
the hidden state that is $(1,0,0)$ if the internal state is $A$, $(0,1,0)$ if
the internal state is $B$, and $(0,0,1)$ if the internal state is $C$.  By
inspection, we have:
\begin{align*}
h_{t+1,A} &= 1-h_{t,A} \\
h_{t+1,B} &= x_t h_{t,A} \\
h_{t+1,C} &= (1-x_t) h_{t,A}
  ~.
\end{align*}
One can straightforwardly find weights that lead to $p_{LSTM}(x_{t+1}|h_t)$
accurately reflecting the transmission (emission) probabilities. In other
words, in theory a three-node RNN (and an equivalent three-node LSTM) can learn
to predict the Neven process optimally.

However, the Neven Process' simplicity is belied by the gap between the
predictors' accuracy and rate and the predictive rate-accuracy curve. The
worst predictive accuracy falls short of the optimal by $\sim15\%$, and none of
the GLMs, RCs, or LSTMs get closer than $\sim 97\%$ to optimal. Furthermore,
almost all the rates surpass the maximal optimal predictor rate.


\subsection{Comparing GLMs, RCs, and LSTMs}

We now analyze the combined results obtained over all minimal PDFAs up to four
states using two metrics. (Again, recall that they are $1,338$ unique
machine topologies.) To compare across PDFAs, we first normalize the rate and
accuracy by the rate and accuracy of the optimal predictor. Then, we find the distance from the
predictor's rate and accuracy to the predictive rate-accuracy curve, which is
similar in spirit to the metric of Ref. \cite{zaslavsky2018efficient} and to
the spirit of Ref. \cite{palmer2015predictive}. 
Note that this metric would
have been markedly harder to estimate had we used nondeterministic probabilistic
finite automata; that is, those without determinism (unifiliarity) in their
transition structure \cite{Marz14f}.

\begin{figure}[!t]
\centering
\includegraphics[width=2.5in]{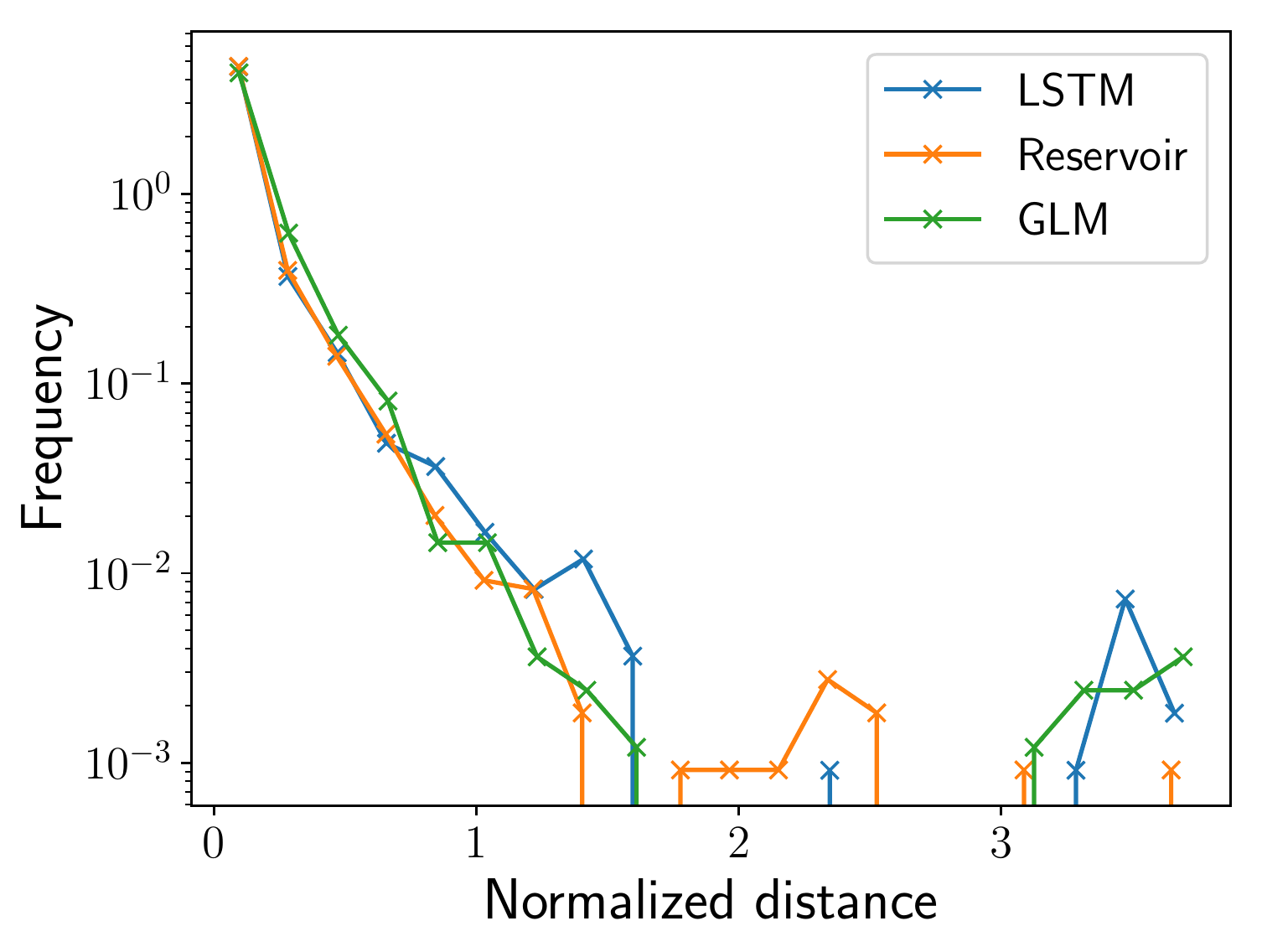}
\caption{Histogram of normalized distances to the predictive rate-accuracy
	curve for LSTMs (blue), RCs (orange), and GLMs (green) using $1755$ distinct PDFAs.
	}
\label{fig:dist_PRD}
\end{figure}


Figure \ref{fig:dist_PRD} showcases a histogram of the normalized distance to the predictive rate-accuracy curve, ignoring PDFAs for which the maximal optimal rate is $0$ nats. 
The normalized distance for all three predictor types tends to be quite small,
but even so, we can see differences in the three predictor types. LSTMs tend to
have smaller normalized distances than RCs, and RCs tend to have smaller
normalized distances to the predictive rate-accuracy curve than GLMs.

The same trend holds for the percentage difference between the predictive
accuracy and the maximal predictive accuracy, which we call the
\emph{normalized predictive distortion}. Trained LSTMs on average have $3.9\%$
predictive distortion; RCs on average have $4.0\%$ predictive distortion; and
GLMs on average have $6.5\%$ predictive distortion. When looking only at
optimized LSTMs, RCs, and GLMs---meaning that the number of nodes or the order
is chosen to minimize normalized predictive distortion---a few PDFAs still have
high normalized predictive distortions of $4.6\%$ for LSTMs, $9.7\%$ for RCs,
and $27.3\%$ for GLMs. Some LSTMs, RCs, and GLMs reach normalized predictive
distortions of as much as $50\%$.
 
Unsurprisingly, increasing the GLM order and the number of nodes of the RCs and
LSTMs tends to increase predictive accuracy and decrease the normalized
distance.

Our final aim is to understand the PDFA characteristics that cause them to be
harder to predict accurately and/or efficiently. We have two suspects, which are the most natural
measures of process ``complexity''. This first is the generated process'
entropy rate $h_{\mu}$, the entropy of the next symbol conditioned on all
previous symbols, which quantifies the intrinsic randomness of the stimulus.
The second is the generated process' statistical complexity $C_{\mu}$, the
entropy of the causal states, which quantifies the intrinsic memory in the
stimulus. The more random a stimulus, the harder it would be to predict;
imagine having to find the optimal predictor for a biased coin whose bias is
quite close to $1/2$. The more memory in a stimulus, the more nodes in a
network or the higher the order of the GLM required, it would seem. We
performed a multivariate linear regression, trying to use $h_{\mu}$ and
$C_{\mu}$ to predict the minimal normalized predictive distortion and minimal
normalized distance. We find a small and positive correlation for
LSTMs, reservoirs, and GLMs for predicting normalized distance, with an $R^2$
of $0.002$, $0.12$, and $0.15$, respectively. We find a larger positive correlation for
LSTMs, reservoirs, and GLMs for predicting normalized distortion, with an $R^2$
of $0.09$, $0.24$, and $0.24$, respectively.  Interestingly, the performance GLMs and RCs is impacted by increased randomness and increased memory in the stimulus, while the LSTMs' accuracy has little correlation with entropy rate and statistical complexity.

\section{Conclusion}

We have known for a long time that reservoirs and RNNs can reproduce any
dynamical system\cite{maass2002real,grigoryeva2018echo,doya1993universality},
and we have explicit examples of RNNs learning to infer the hidden states of a
PDFA when shown the PDFA's output \cite{cleeremans1989finite}. We revisited
these examples to better understand if the finding of Ref.
\cite{cleeremans1989finite} is typical. How often do RNNs and RCs learn
efficient and accurate predictors of PDFAs, especially given that BSI can yield
an optimal predictor with orders-of-magnitude less training data?

We conducted a rather comprehensive search, analyzing all $1,388$
randomly-generated PDFAs with four states or less. For each PDFA, we trained
GLMs, RCs, and RNNs of varying orders or varying numbers of nodes. Larger
orders and larger numbers of nodes led to more accurate and more efficient
predictors. On average, the various time series predictors have $\sim 5\%$
predictive distortion. In other words, we are apparently better at classifying
MNIST digits than sometimes predicting the output of a simple PDFA. And again,
existing algorithms \cite{Stre13a} can optimally predict the output of the
PDFAs considered here with orders-of-magnitude less training data. These
findings lead us to conclude that algorithms that explicitly focus on inference
of causal states \cite{Crut88a, Stre13a, pfau2010probabilistic,
still2014information} have a place in the currently RNN-dominated field of time
series prediction.


\section*{Acknowledgment}

This material is based upon work supported by, or in part by, the Air Force
Office of Scientific Research under award number FA9550-19-1-0411 and the U. S.
Army Research Laboratory and the U. S. Army Research Office under contracts
W911NF-13-1-0390 and W911NF-18-1-0028.

\ifCLASSOPTIONcaptionsoff
  \newpage
\fi



\bibliographystyle{IEEEtran}
\bibliography{chaos}


\end{document}